\documentclass[preprint,12pt]{elsarticle}

\usepackage{epsfig}
\usepackage{amssymb}
\usepackage{kbordermatrix}

\journal{}

\newcommand{\argmin}{\arg\!\min}
\newcommand*{\bigchi}{\mbox{\Large$\chi$}}

\begin{document}

\begin{frontmatter}



\title{A global approach for learning sparse Ising models.}


\author{Daniela De Canditiis}

\address{Istituto per le Applicazioni del Calcolo CNR,  Rome Italy}

\begin{abstract}
We consider the problem of learning the link parameters as well as the structure of a binary-valued pairwise Markov model. Under sparsity assumption, we propose a method based on $l_1$- regularized logistic regression, which estimate globally the whole set of edges  and link parameters. Unlike the more recent methods discussed in literature that learn the edges and the corresponding link parameters one node at a time, in this work we propose a method that learns all the edges and corresponding link parameters simultaneously for all nodes. The idea behind this proposal is to exploit  the reciprocal information of the nodes between each other during the estimation process. Numerical experiments highlight the advantage of this technique and confirm the intuition behind it.  

\end{abstract}


\end{frontmatter}

\section{Introduction}
\label{sec:intro}
Ising models are fundamental undirected binary graphical models that capture pairwise dependencies
between the input variables. They are well-studied in the literature and have applications
in a large number of areas such as physics, computer vision and statistics (\cite{Jaimovich et al. (2006)}; \cite{Koller et al. (2009)};
\cite{Vuffray et al. (2016)}). One of the core problems
in understanding graphical models is structure learning, that is, recovering the structure of the
underlying graph given a random sample from the population distribution. The literature is very rich and it is difficult to make a complete picture of this topic. In the seminal paper, \cite{Chow1968} proposed a greedy algorithm for undirected graphical models assuming the underlying graph is a tree. Since then, various algorithms
have been proposed for structure estimation of graphical models.
They can be broadly classified into two categories: combinatorial algorithms
 and convex relaxation algorithms. The first category of methods  are based on an empirical estimate of some statistics by which performing local tests on small groups of data, and
then combining them to output a graph structure: among many others \cite{Bresler et al.2008} and \cite{Bresler and Guy2015} ultimately propose an algorithm based on the empirical frequencies, \cite{Anandkumar et al. 2012}  propose an algorithm based on empirical conditional variation distance, and \cite{Klivans et al. 2017}  proposes a sparsitron algorithm based on a weighted voting scheme. All these methods focus on graph estimation but not on parameters estimation.  On the other hand, the second category  includes methods based on penalized convex optimization. Among many others, under different assumptions on the structure of the underlying graph \cite{Lee et al. (2007)}, \cite{Yang et al. (2016)} and  \cite{Ravikumar et al. 2010}. In particular, \cite{Ravikumar et al. 2010} was a pioneering paper in this direction proposing to learn the graphical structure by a \emph{node-wise} approach using an $\ell_1$-penalized logistic regression. In that paper for the first time the authors were also interested in estimating the link parameters, or more precisely their sign. 
 At the same time, in \cite{Hofling and Tibshirani 2009}, exploiting the same idea, attention was paid to both aspects, namely the estimation of the graph structure  as well as the estimation of the values of the link parameters and not only their sign. 

We start from these last two works and propose a modification of the method in  \cite{Ravikumar et al. 2010}, that allows us to estimate the structure of the graph and the link parameters (not only their sign) in a global manner and not one node at a time. The proposed global approach takes the form of a parallelization of the node-wise approach, in the sense that it learns all the nodes simultaneously, thus exploiting the mutual information of each node on the others.

The paper is organized as follows. In Section \ref{sec: math} we present the mathematical and statistical formulation of the problem. In Section \ref{sec: methods} we present the node-wise approach which represents the starting point of the proposed method, and in the same section we describe our global approach. 
In the last Section, we show a simulation study comparing the node-wise and the global approach confirming the intuition behind it.

\section{Mathematical framework: the Ising model} \label{sec: math}
For a complete and exhaustive treatment of graphs theory we refer to \cite{Lauritzen}; below we give only definitions and properties necessary for this work.
A finite graph $G=(V,E)$ consists of a finite collection of \textit{nodes} $V=\{1,2...,p\}$  and a collection of \textit{edges} $E \subseteq V\times V$. For the scope of this work, we  consider graphs that are \textit{undirected}, namely graphs whose edges are not ordered, i.e. there is no distinction between the edges $(i,j)$ and $(j,i) \in E$. Moreover, for any $i \in V$, $N(i) := \{j \in V : (i,j) \in E\}$ is the set of neighbours of node $i$ and $C \subset V$ is a \textit{clique} if $(i,j)\in E$ for all $i,j \in C$ such that $i \neq j$.  

In this paper the notion of a graph is used to keep track of the conditional dependence relationship between random variables of a complex system.
By complex system here we mean a jointly distributed random variables $(X_1,X_2,...,X_p)$ that interact with each other. In the following, a formal definition of conditional independence relationship is given:
\vspace{0.4cm}

{\bf Definition:}\label{def1}
  Two random variables ($X_i$, $X_j$) of a random vector $(X_1, \ldots,X_p)$ are conditionally independent, $ X_i  \perp X_j | X_{V\backslash\{i,j\}}$, if

\begin{equation}
\begin{array}{c}
\quad f(x_i,x_j | x_{V\backslash\{i,j\}})=f(x_i | x_{V\backslash\{i,j\}}) f(x_j | x_{V\backslash\{i,j\}}) \\
\Updownarrow \\
f(x_i | x_{V\backslash\{i\}}) \mbox{   does not depend on   } x_j
\end{array}
\end{equation}

where $f(\cdot)$ stands for density distribution or probability mass function  and $x_{S}:=(x_s, s \in S)$, with $S$ any subspace of $V$.

\vspace{0.4cm}

Associated with an undirected graph $G=(V,E)$ and a system of random variables $X_V$ indexed in the vertexes set $V$ there is a range of different Markov properties which establish how much the graph is explanatory of the conditional independence property of the random variables, see \cite{Lauritzen}. Specifically, in this work we deal with system of random variables which are pairwise Markov with respect to an undirected graph $G=(V,E)$, so it holds that
$$
X_i  \perp X_j | X_{V\backslash\{i,j\}} \quad \Leftrightarrow \quad (i,j) \notin E,
$$  
which establishes conditional independence among two variables $X_i$ and $X_j$ if and only if their corresponding nodes in the graph $G$ are not connected.
Another way an undirected graphical model can encode the conditional dependency relationships
between the system variables is through the factorization property. Let $\mathcal{C}$ be  the
set of all possible cliques in a graph $G$, then the distribution factorizes as
\begin{equation}
f(x_1,...,x_p)= \prod_{C \in \mathcal{C}} \phi(x_C),
\end{equation}
where $\phi(X_C)$ is a positive function ({\emph{potential function}) which depends only on the variables corresponding to the nodes in $C$, $\forall C \in \mathcal{C}$.
Under the hypothesis that the joint distribution $f(x_1,...,x_p)$ is positive, the pairwise Markov property and the factorization property are equivalent as claimed by the Hammersley-Clifford theorem proved in (\cite{Lauritzen} cfr. Theorem 3.9).

In this paper, we work under such hypothesis, in particular with a system of binary random variables with values in $\{ -1,1\}$ and for which the multivariate distribution (probability mass function in such a case) factorizes in the following way:

\begin{equation}  \label{eq: model_Ising_general}
P(x_1,\ldots,x_p)= \frac{1}{Z(\theta)} exp \left( \sum_{i=1}^p \theta_i x_i + \sum_{(i,j) \in E} ~ \theta_{ij}x_i x_j \right)
\end{equation}
with  $\theta_i$ and $\theta_{ij} \in \mathbb{R}$ some parameters and  $Z(\theta)$ a constant making the probabilities sum to one (usually called partition function), see \cite{Lauritzen}.
The set $E$ represents the edges' set of the undirected graph which is pairwise Markov with respect to the distribution and it represents the conditional dependency relationships among the system variables.
This model is known as {\emph{Ising model}} and it is used in many application of spatial statistics such as modelling the behaviour of ferromagnets, 
since in such case the discrete variables represent magnetic dipole moments of atomic spins arranged in a graph 
that can be in one of two state $\{ +1,-1\}$, see \cite{Ravikumar et al. 2010}. 
In particular, in this paper we consider model with no first order terms, 
which corresponds to have a zero external field interacting with the ferromagnets' system:
\begin{equation}  \label{eq: model_Ising}
P(x_1,\ldots,x_p)= \frac{1}{Z(\theta)} exp \left( \sum_{(i,j) \in E} ~ \theta_{ij}x_i x_j \right).
\end{equation} 
From a probabilistic point of view, having no first order terms makes the model symmetric to switching the value of the variable in all the graph nodes, i.e. $P(x_1,\ldots,x_p)=P(-x_1,\ldots,-x_p)$. This is the model considered in \citep{Ravikumar et al. 2010}.

Note that, since in  graph $G=(V,E)$ there is no distinction between edge $(i,j)$ and $(j,i)$, in Equation (\ref{eq: model_Ising}) there is no distinction between parameter $\theta_{ij}$ and $\theta_{ji}$, from a physical point of view this mean that the  interaction strength between two variables is a reciprocal/symmetric relationship. For convenience of exposition, we define the vector $x=(x_1,\ldots,x_p)$ and the symmetric zero diagonal matrix $\Theta$ by which it is possible to express the joint probability in (\ref{eq: model_Ising}) by the following formula 
$$
P(x)= \frac{1}{Z(\theta)} exp \left( x^t \Theta x /2 \right).
$$ 
Matrix $\Theta$ is our unknown parameter; its support encoding the edges' set $ E $ of the  graph, its $ p (p-1) / 2 $ upper off-diagonal elements representing the link parameters.

Our perspective is inferential. Given a sample of size $n$ extracted from an unknown distribution $P(x)$ of form (\ref{eq: model_Ising}), as the one represented in Table \ref{tab_sample}, we want to learn both the structure of the undirected graph $G$ which is Markov with respect to the distribution as well as the link parameters itself. In other words, we want to learn the dependence/independence conditional relationships between the system variables (i.e. the pairwise Markov graph structure) as well as the strength of these relationships (i.e. the numerical value of link parameters $\theta_{ij}$). Both of these objectives are achieved if matrix $\Theta$ is correctly estimated, i.e. if the whole vector $(\theta_{ij})_{i<j}$ of its upper off-diagonal elements is correctly estimated.

\begin{table}
\caption{Example of a sample of size $n$ from a generic distribution $P(x_1,x_2,\ldots,x_p) $}
\label{tab_sample}
\begin{center}\scriptsize
\begin{tabular}{ c | c c c c }
&  $X_1$ &    $X_2$   &  \ldots &   $X_p$ \\
\hline
$x^{(1)}$ & $-1$ & $+1$ & \ldots & $+1$ \\
$x^{(2)}$ & $+1$ & $-1$ & \ldots & $-1$ \\
$x^{(3)}$ & $-1$ & $-1$ & \ldots & $+1$ \\
$x^{(4)}$ & $-1$ & $-1$ & \ldots & $+1$ \\
$x^{(5)}$ & $-1$ & $-1$ & \ldots & $-1$ \\
          & \vdots & \vdots & \vdots & \vdots \\
$x^{(n)}$ & $-1$ & $-1$ & \ldots & $+1$ \\
\end{tabular}
\end{center}
\end{table}

\section{Estimation methods} \label{sec: methods}

In the milestone paper \cite{Ravikumar et al. 2010} the authors propose one of the most efficient methods for recovering the graph $G$ associated to an Ising model as well as the sign of its link parameters, $sign(\theta_{ij})$. Their method does not make hypotheses on the graph's structure apart from the sparsity. From that work many others have emerged in the literature, lastly \cite{Wu et al. (2018)}. In the following we revise it to make our proposal clearer.

Let us consider the conditional distribution of one system  variable with respect to the others. Let us fix a variable $X_r$, with $r \in \{1,2,..p\}$ and consider the set of the remaining  variables $X_{V \setminus \{r\}}$; define $E_{-r}=E \setminus \{ (i,r): i \in N(r) \}$ as the subset of edges not involving the $r$-th node, hence we have
$$
\begin{array}{ccl}
P_{\theta_{\cdot r}}(x_r | x_{V \setminus \{r\}})&=&\frac{P(x_V)}{P(x_{V \setminus \{r\}})} =\frac{P(x_1,\ldots,x_p)}{P(x_1,\ldots,x_{r-1},1,x_{r+1}\ldots,x_p)+ P(x_1,\ldots,x_{r-1},-1,x_{r+1}\ldots,x_p)} \\
\\
 &=&\frac{ exp \left( \sum_{i \in N(r)} ~ \theta_{ir}x_i x_r+  \sum_{(i,j) \in E_{-r}} ~ \theta_{ij}x_i x_j \right)}{exp \left(\sum_{i \in N(r)} ~ \theta_{ir}x_i+  \sum_{(i,j) \in E_{-r}} ~ \theta_{ij}x_i x_j \right) +
 exp \left(-\sum_{i \in N(r)} ~ \theta_{ir}x_i+ \sum_{(i,j) \in E_{-r}} ~ \theta_{ij}x_i x_j \right)}\\
\\
 &=& \frac{ exp \left(  2 x_r \sum_{i \in N(r)} ~ \theta_{ir}x_i   \right)}{exp \left(2 x_r  \sum_{i \in N(r)} ~ \theta_{ir}x_i \right) +1}.
 \end{array} 
$$ 

Evaluating the previous formula for $x_r=1$, it follows that
$$
P(X_r=1 |x_{V \setminus \{r\}})=\frac{ exp \left( \sum_{i \in N(r)} ~ 2\theta_{ir}x_i   \right)}{exp \left(\sum_{i \in N(r)} ~ 2\theta_{ir}x_i \right) +1}= logistic \left( \sum_{i \in N(r)} ~ 2\theta_{ir}x_i \right),
$$
where $logistic(*)=e^*/(e^*+1)$. 
It is then possible to learn  vector $\theta_{\cdot r}$ solving a $l_1$-penalized logistic regression problem, where $X_r$ and $X_{V \setminus \{r\}}$ play the role of response variable  and covariates respectively. 

In particular, let $\{ x^{(i)} \in \{ +1,-1\}^p \}_{i=1,...,n}$ be the sample of size $n$ extracted from the unknown population, then the loglikelihood function for the unknown parameter vector $\theta_{\cdot r}=(\theta_{lr})_{l \neq r}$ is the following function
$$
\ell (\theta_{\cdot r}):= \sum_{i=1}^n log P_{\theta}(x^{(i)}_r | x^{(i)}_{V \setminus \{r\}})
$$
with 
\begin{equation} \label{eq: likelihood}
P_{\theta}(x_r | x_{V \setminus \{r\}})=
\frac{exp \left(x_r \sum_{l \neq r} 2 \theta_{lr} x_l\right) }{exp \left( x_r \sum_{l \neq r} 2 \theta_{lr} x_l\right) +1}.
\end{equation}
Since we expect that the elements of the unknown vector corresponding to the neighbours of  node $ r $ are different from zero (i.e. $\theta_{lr} \neq 0 $ $\leftrightarrow$ $l \in N(r)$), while the elements corresponding to the non-neighbours of  node $ r $  are zero (i.e. $\theta_{lr} = 0 $  $\leftrightarrow$  $l \notin N(r)$), the following $l_1$-penalized logistic regression problem is solved
\begin{equation} \label{eq:unalogistic}
\hat{\theta}_{\cdot r}=\argmin_{\theta \in \mathcal{R}^{p-1}} -\frac{1}{n} \sum_{i=1}^n log P_{\theta}(x^{(i)}_r | x^{(i)}_{V \setminus \{r\}}) + \lambda \| \theta\|_1;
\end{equation}
 $N(r)$ being the true set of  active variables into problem in Eq. (\ref{eq:unalogistic}) with cardinality $d_r=card(N(r))$ equals the degree of node $r$.
Let us revise the theoretical property of this estimator, as exposed in the original paper \cite{Ravikumar et al. 2010}. Denote the true model parameter $\theta^{\star}$. For any fixed node $r \in V$, define the Hessian  matrix ($p-1 \times p-1$) of the likelihood function $\mathbb{E}(log(P_{\theta}(X_r|X_{V \setminus \{r\}}))$ as evaluated at the true model parameter $\theta^{\star}_{\cdot r}$

\begin{equation}
Q^{\star}_r:=\mathbb{E}_{\theta^{\star}_{\cdot r} }\left\{ \nabla^2 log( P_{\theta}(X_r|X_{V \setminus \{r\}} ) \right\},
\end{equation}
and denote $Q^{\star}_1$ and $Q^{\star}_2$ the restriction  of $Q^{\star}_r$ to the index sets $N(r) \times N(r)$ and  $V \backslash \{N(r)  \cup \{r\}\} \times N(r)$ respectively, i.e.  $Q^{\star}_1$ and $Q^{\star}_2$ have dimension $d_r \times d_r$  and $p-d_r-1 \times d_r$, respectively. We can now state the following three assumptions:
\begin{enumerate}
\item[A1)] \emph{Dependency condition}: there exists a constant $C_{min}$  such that
$$
\Lambda_{min}(Q^{\star}_1) \geq C_{min},
$$
with $\Lambda_{min}$ the minimum eigenvalue. 
\item[A2)] \emph{Incoherence condition}: there exist $\alpha \in (0,1]$ such that
$$
\| Q^{\star}_2 (Q^{\star}_1)^{-1}\|_{\infty} \leq 1-\alpha.
$$
\item[A3)] \emph{Minimum strength signal}: 
$$
\theta^{\star}_{min}:=min_{i \neq r} |\theta^{\star}_{ir}| \geq \frac{10}{C_{min}} \sqrt{d_r} \lambda
$$
\end{enumerate}

Under assumptions A1), A2) and A3), Theorem 1 in \cite{Ravikumar et al. 2010} proves that, when $\lambda=\mathcal{O}(\sqrt{log p/n})$ and $n$ is big enough, with high probability the solution $\hat{\theta}_{\cdot r}$ of problem (\ref{eq:unalogistic}) consistently estimates the signed set of node $r$ neighborhood, i.e. 
$$
\hat{N}_{\pm}(r)={N}_{\pm}(r),
$$
with
\begin{eqnarray*}
\hat{N}_{\pm}(r):=\{ sign( \hat{\theta}_{ir})i: i \in V \setminus \{r\} , \hat{\theta}_{ir} \neq 0\} \\
{N}_{\pm}(r):=\{ sign( \theta^{\star}_{ir})i: i \in V \setminus \{r\} , \theta^{\star}_{ir} \neq 0\}.
\end{eqnarray*}

The reason why the authors do not consider $\hat{\theta}_{ir}$ as estimate of the link parameter lies in the fact that they solve problem (\ref{eq:unalogistic}) for each node $ r \in V $ independently of the other nodes so that $ \hat{\theta}_{ir} \neq \hat{\theta}_{ri}$. This means that, if for each variable $X_r$ we place the estimate of its link parameter vector  $\hat{\theta}_{\cdot r}$ into the extra diagonal elements of the corresponding column of matrix $\hat{\Theta}$, then this leads to a non-symmetric estimated matrix. For this reason in  \cite{Hofling and Tibshirani 2009} two procedures for symmetrizing this method are proposed. 
The first procedure, named N-L-m ({\emph{Neighborhood-based Logistic minimum}}), works in the following way:
\begin{equation} \label{eq: N-L-m}
\hat{\theta}^{N-L-m}_{ij}= \hat{\theta}^{N-L-m}_{ji}=\left\{ \begin{array}{cl}
\hat{\theta}_{ij} & if |\hat{\theta}_{ij}| > |\hat{\theta}_{ij}| \\
\hat{\theta}_{ji} & if |\hat{\theta}_{ij}| \leq |\hat{\theta}_{ij}|
\end{array}   \right. ,
\end{equation}
similarly, the second procedure, named N-L-M ({\emph{Neighborhood-based Logistic Maximum}}), works in the following way:
\begin{equation} \label{eq: N-L-M}
\hat{\theta}^{N-L-M}_{ij}=\hat{\theta}^{N-L-M}_{ji}= \left\{ \begin{array}{cl}
\hat{\theta}_{ij} & if |\hat{\theta}_{ij}| < |\hat{\theta}_{ij}| \\
\hat{\theta}_{ji} & if |\hat{\theta}_{ij}| \geq |\hat{\theta}_{ij}|
\end{array}   \right. .
\end{equation}

We stress that, both procedures consist of two steps:
in a first step for each node $r \in V$  vector $\hat{\theta}_{\cdot r}$ is estimated by solving (\ref{eq:unalogistic}), in a second step procedure (\ref{eq: N-L-m}) and (\ref{eq: N-L-M}) is applied respectively. 
 
\subsection{Global Logistic method}
  
In this paper we propose a new method for learning both the graph $G=(V,E)$ as well as the link parameters $(\theta_{ij})_{i<j}$. The idea behind our proposal is to learn matrix $\Theta$ globally instead of column-wise as in the previous two methods. Our idea can be seen as a kind of parallelization of the node-wise approach, in the sense that we learn all the nodes simultaneously, thus using the reciprocal information of each node on the others. In fact, when solving problem (\ref{eq:unalogistic}) for node $r$, parameter $ \theta_{ir} $ also come into play when solving problem (\ref{eq:unalogistic}) for node $i$. This fact is more clear if we write the $p$ independent problems in one system of problems:

\begin{equation} \label{eq:system}
\left\{ \begin{array}{c}
\argmin -\frac{1}{n} \sum_{i=1}^n log P_{\theta_{\cdot 1}}(x^{(i)}_1 | x^{(i)}_{V \setminus \{1\}}) + \lambda_1 \sum_{i \neq 1} |\theta_{i1}| \\
\\
\argmin -\frac{1}{n} \sum_{i=1}^n log P_{\theta_{\cdot 2}}(x^{(i)}_2 | x^{(i)}_{V \setminus \{2\}}) + \lambda_2 \sum_{i \neq 2} |\theta_{i2}| \\
\vdots \\
\argmin -\frac{1}{n} \sum_{i=1}^n log P_{\theta_{\cdot p}}(x^{(i)}_p | x^{(i)}_{V \setminus \{p\}}) + \lambda_p \sum_{i \neq p} |\theta_{ip}| \\
\end{array} \right.
\end{equation}
We stress that the solution of system (\ref{eq:system}) are $p$ vectors each of length $p-1$, hence a total of $p(p-1)$ parameters are obtained instead of $p(p-1)/2$ of real interest. Our proposal is to solve the following  problem:
\begin{equation} \label{eq:G-L}
\hat{\theta}=\argmin -\frac{1}{np} \sum_{r=1}^p \sum_{i=1}^n log P_{\theta_{\cdot r}}(x^{(i)}_r | x^{(i)}_{V \setminus \{r\}}) + \lambda \sum_{i < j} |\theta_{ij}|
\end{equation}

where $ P_{\theta_{\cdot r}}(x^{(i)}_r | x^{(i)}_{V \setminus \{r\}})$ is given in eq. (\ref{eq: likelihood}) and the number of unknown parameters is exactly $p(p-1)/2$. The same idea of jointly infer the neighbourhood of all the graph nodes simultaneously has been proposed in \cite{Friedman et al 2000} for the problem of learning Gaussian Graphical Models. To highlight the fact that this method learns the graph in a global fashion and not node-wise, we call this procedure  {\emph{Global Logistic}} (G-L). We can now state a theoretical property of this estimator, by exploring results for a single node logistic regression in \cite{Ravikumar et al. 2010}. Define the Hessian  matrix ($p(p-1)/2 \times p(p-1)/2$) of the likelihood function $\mathbb{E}( log( P_{\theta}(X)))=\mathbb{E}(\sum_{r=1}^p log(P_{\theta_{\cdot r}}(X_r|X_{V \setminus \{r\}}))$ as evaluated at the true model parameter $\theta^{\star}:=(\theta^{\star}_{ij})_{i<j}$ 

\begin{equation}
Q^{\star}:=\mathbb{E}_{\theta^{\star} }\left\{ \nabla^2 log( P_{\theta}(X) \right\}.
\end{equation}
Define the row/column index set $S:=\{ij : (i,j) \in E \}$, note that it equals the set of active variables for problem in Eq. (\ref{eq:G-L}), i.e. $S=\{ij : \theta^{\star}_{ij} \neq 0 \}$ and, moreover, define $Q^{\star}_1$ and $Q^{\star}_2$ the restriction  of $Q^{\star}$ to $S \times S$ and  $S^c  \times S$ respectively, i.e.  $Q^{\star}_1$ and $Q^{\star}_2$ have dimension $d \times d$  and $p(p-1)/2-d \times d$, respectively with $d=card(S)$. We can now state the following three assumptions, similar to those in the previous section :
\begin{enumerate}
\item[B1)] \emph{Dependency condition}: there exists a constants $C_{min}$ such that
$$
\Lambda_{min}(Q^{\star}_1) \geq C_{min} 
$$

\item[B2)] \emph{Incoherence condition}: there exist $\alpha \in (0,1]$ such that
$$
\| Q^{\star}_2 (Q^{\star}_1)^{-1}\|_{\infty} \leq 1-\alpha.
$$
\item[B3)] \emph{Minimum strength signal}: 
$$
\theta^{\star}_{min}:=min_{i < j} |\theta^{\star}_{ij}| \geq \frac{10}{C_{min}} \sqrt{d} \lambda.
$$
\end{enumerate}

Under assumptions B1), B2) and B3), by using Theorem 1 in \cite{Ravikumar et al. 2010}, we can state that when $\lambda=\mathcal{O}(\sqrt{log (p(p-1)/2) /np})$ and $n$ is big enough, with high probability the solution $\hat{\theta}$ of problem (\ref{eq:G-L}) consistently estimates the set of active variables, i.e. $\hat{S}=S$
with
\begin{eqnarray*}
\hat{S}:=\{ij : \hat{\theta}_{ij} \neq 0\}.\\
\end{eqnarray*}
This result ensures the consistency of the method at least in terms of support of the matrix $\Theta$, but its assumptions are neither verifiable nor usable from a practical point of view. It remains however the possibility to use the choice of $\lambda$ suggested by this result, even if not optimal, to make a first estimate of the matrix $\Theta$. This is, in fact, the choice of $\lambda$ made by the same authors in \cite{Ravikumar et al. 2010} to perform their numerical tests, and therefore, this is the choice that we make to perform our experiments in the next Section. In the following Section, in fact,  we show some numerical tests to understand how the proposed procedure improves with respect to the N-L-m and N-L-M presented above. 
Finally, it is worthwhile to observe that, in principle from computational point of view, there is almost no difference in solving $p$ independent $l_1$-penalized logistic regression problems of size $n$ and dimension $p-1$ each, and solving one single $l_1$-penalized logistic regression problem with size $np$ and dimension $p(p-1)/2$; however this can be not properly true if a standard software package  is used to solve the optimization problem, without exploiting the particular structure of the  likelihood function. On the other hand, there is difference in selecting $p$ different regularization parameters ($\lambda_1, \ldots, \lambda_p$) when applying N-L-m and N-L-M procedures, and selecting one single parameter $\lambda$ when applying G-L procedure. Hence, in principle, this is an important advantage of the global approach with respect to the node-wise approach.

\section{Numerical experiments}

In this Section, we show some numerical experiments to study the performance of the proposed method. 
Before presenting results it is necessary to specify how we generated data, what algorithm we used, how we chose the regularization parameter and what kind of indexes we used to measure performance.

\vspace{0.4cm}

\emph{Data generation:}
let us denote $\bigchi$ as data matrix of dimension $n \times p$ which represents $n$ independent realizations of an Ising model of type (\ref{eq: model_Ising}). This means that we fix a matrix $\Theta$ (to be specified later) and we generate samples from distribution given in (\ref{eq: model_Ising}). While in the case $p \leq 10 $ this is possible for any computer resource, in the case $p>10$ the set of possible realizations of system variables, $2^p$, may became prohibitive. For this reason, we describe the Gibbs sampling scheme we have adopted to generate samples from distribution (\ref{eq: model_Ising}) when its dimension  $p$ is high.

\begin{itemize}

\item \emph{randomly choose an initial state:}  
\begin{itemize}

\item[] $x^{(0)} \in \{ +1,1\}^p$ 

\end{itemize}

\item \emph{for k=1,... generate a new sample $x^{(k)}$  by the following procedure:}
\begin{itemize}
\item[] for $r=1:p$

\item[] sample $x_r^{(k)}$ from $ P(x^{(k)}_r | x^{(k)}_{1},...,x^{(k)}_{r-1},x^{(k-1)}_{r+1},...,x^{(k-1)}_{p})$ given in (\ref{eq: likelihood})
  
\item[] end
\end{itemize}
\end{itemize}                   

of course the $n$ samples are taken after an appropriate burn-in time. 

\vspace{0.4cm}

\emph{Algorithm:}
we use the Matlab function {\emph{lassoglm}} to solve both problem (\ref{eq:system}) and problem (\ref{eq:G-L}), see \cite{lassoglm} for details. While for solving problem (\ref{eq:unalogistic}) we use the $r$-th column of data matrix $\bigchi_{\cdot r}$ as regressor and the remaining columns as covariates, for solving problem (\ref{eq:G-L}) the construction of regressor and covariates is more complicated. For completeness we describe it in the simplest case $p=3$ for which we have:
$$
\begin{array}{c}
P(X_1=1 | x_2 , x_3)= {exp(\theta_{12} x_2 + \theta_{13} x_3)} / {(exp(\theta_{12} x_2 + \theta_{13} x_3) +1)} \\
P(X_2=1 | x_2 , x_3)= {exp(\theta_{12} x_1 + \theta_{23} x_3)}/ {(exp(\theta_{12} x_1 + \theta_{23} x_3) +1)} \\
P(X_3=1 | x_1 , x_2)= {exp(\theta_{13} x_1 + \theta_{23} x_2)}/ {(exp(\theta_{13} x_1 + \theta_{23} x_2) +1)}. \\
\end{array}
$$
In this case the regressor vector, the covariates matrix and the unknown parameters vector are the following:
$$
\kbordermatrix{& y \cr
                  &\bigchi_{\cdot 1}  \cr
                  &\bigchi_{\cdot 2}   \cr
                  &\bigchi_{\cdot 3}  \cr
               }
                \quad 
\kbordermatrix{&\tilde{X}_1 &\tilde{X}_2 &\tilde{X}_3 \cr
                &\bigchi_{\cdot 2}  & \bigchi_{\cdot 3}  & 0 \cr
                 &\bigchi_{\cdot 1}  & 0 & \bigchi_{\cdot 3}  \cr
                 &0 & \bigchi_{\cdot 1}  & \bigchi_{\cdot 2} \cr
                }
               \quad
               \kbordermatrix{& \theta \cr
                  &\theta_{12}  \cr
                  &\theta_{13}    \cr
                  &\theta_{23}   \cr
               },                 
$$
being $y$ a sample of size $3n$ of a  binary random variable $Y$ and $p(p-1)/2=3$ the effective number of unknown link parameters. With this choice indeed we have that 
$$
\begin{array}{cl}
P(Y=1| \tilde{x_1}, \tilde{x_2} ,\tilde{x_3}) & =logistic(\theta_{12} \tilde{x_1}+\theta_{13} \tilde{x_2}+\theta_{23} \tilde{x_3}) \\
& = exp(\theta_{12} \tilde{x_1}+\theta_{13} \tilde{x_2}+\theta_{23} \tilde{x_3}) / (exp(\theta_{12} \tilde{x_1}+\theta_{13} \tilde{x_2}+\theta_{23} \tilde{x_3}) +1)
\end{array}.
$$

\vspace{0.4cm}

\emph{Choice of $\lambda$:}
in order to be fair in a comparative study it is legitimate to fix regularization parameter $\lambda =\sqrt{log (p) / n}$, with $p$ number of unknown parameters and $n$ sample size, which is known from the results revised in Sections \ref{sec: methods} to be order of the optimal regularization parameter. The same choice is indeed done in the numerical experiments of both \cite{Ravikumar et al. 2010} and \cite{Hofling and Tibshirani 2009} respectively. 
Hence, in our experiments in each logistic regression of (\ref{eq:system}) we fix  $\lambda=\sqrt{log(p-1)/n}$,
while in logistic regression (\ref{eq:G-L}) we fix $\lambda=\sqrt{log(p(p-1)/2)/pn}$. Of course, this is not the optimal choice of $\lambda$, which should be performed by a model selection procedure like, CV (Cross Validation), BIC (Bayesian Information Criterion) or MDL (Minimum Description Lenght); however, such a problem  is not pursued here due its computational demanding effort.

\vspace{0.4cm}

\emph{Performance indexes:}
since we are interested both in reconstructing the structure of the graph and in estimating the link parameters, we calculate two different indexes of performance.
The first index measures how the method correctly estimates the structure of the graph and it is defined as:
\begin{equation}  \label{eq:accuracy}
Accuracy= (TP + TN) / (TP + TN + FN + FP),
\end{equation}
where \begin{itemize}
\item[]$TP$ is the number of edges present in the graph and correctly identified (i.e. $\theta_{ij} \neq 0 \wedge \hat{\theta}_{ij} \neq 0$),
\item[]$TN$ is the number of edges not present in the graph and correctly identified  (i.e. $\theta_{ij} = 0 \wedge  \hat{\theta}_{ij}=0$),
\item[]$FN$ is the number of edges present in the graph and not correctly identified (i.e. $\theta_{ij} \neq 0 \wedge  \hat{\theta}_{ij} =0$) and
\item[]$FP$ is the number of edges not present in the graph and not correctly identified  (i.e. $\theta_{ij} =0\wedge  \hat{\theta}_{ij} \neq 0$).
\end{itemize}
In all previous definitions $\hat{\theta}$ is the estimator obtained in Eqs (\ref{eq: N-L-m}), (\ref{eq: N-L-M}) and (\ref{eq:G-L}) respectively.
Note that the measure in (\ref{eq:accuracy}) is a scaled measure inherited from the binary classification literature, $0 \leq accuracy \leq 1$.

\noindent
The second index measures how the method correctly estimates the link parameters and it is defined as the $\ell_2$-norm of the difference between the true and estimated parameters vector:
\begin{equation}
Err= \sum_{i<j} (\theta_{ij}-\hat{\theta}_{ij})^2.
\end{equation}

We can now describe the specific settings we chose for numerical experiments. We generated matrix $\Theta$ according to the \emph{mixed coupling} model, meaning that $\theta^{\star}_{ij}=\pm 0.5$ with equal probability as described in Section 6 of \cite{Ravikumar et al. 2010}.  We propose three examples, meaning that we generated three matrices $\Theta$, of different sizes ($p=25, 50, 100$) and different sparsity regimes, specifically 12\%, 6\% and 3\% are the percentages of non-zero elements respectively. For each of the three examples we consider four different sample sizes mimicking different dimensional regimes. 
In Figures \ref{fig:perfG25}, \ref{fig:perfG50} and \ref{fig:perfG100} we plot the $accuracy$ obtained on 20 independent runs by all the three methods N-L-m, N-L-M ,G-L for the three examples respectively. We can observe that, as expected,  $accuracy$ improves when the sample size increases for all the three methods, however the mutual proportions between the $accuracy$ of the different methods preserve in almost all the cases.  It is important to stress that, accuracy is not optimal in such numerical experiments since the choice of $\lambda$ is fixed a priori for all the methods; to improve $accuracy$ a proper model selection technique should be implemented, but for comparison purpose this is not necessary. The same choice of $\lambda$ was made in \cite{Ravikumar et al. 2010}.     
We can observe that in terms of accuracy the G-L approach slightly improves with respect to N-L-m and N-L-M, but this is not surprising since the methods work under the same hypothesis and almost with the same estimator. Similar observations can be drown, with respect to the estimation  error, i.e. the ability of estimator in reconstructing the value of the link parameters. In Figures  \ref{fig:est_errorG25},  \ref{fig:est_errorG50} and \ref{fig:est_errorG100} we plot the corresponding $Err$ index obtained on the same independent runs and settings. The advantage of G-L with respect to N-L-m and N-L-M is now more evident across all the examples and across all the sample sizes.  The ability of the G-L estimator to better estimate the link parameters is due to the better use of the information, because it focuses on the exact number of unknown parameters. This also explains why, increasing the sample size, this difference  decreases: as the number of data increases, information increases so that the advantage of G-L with respect to N-L-m and N-L-M is less evident.

\begin{figure} 
\centering{
 \includegraphics[width=0.75\columnwidth]{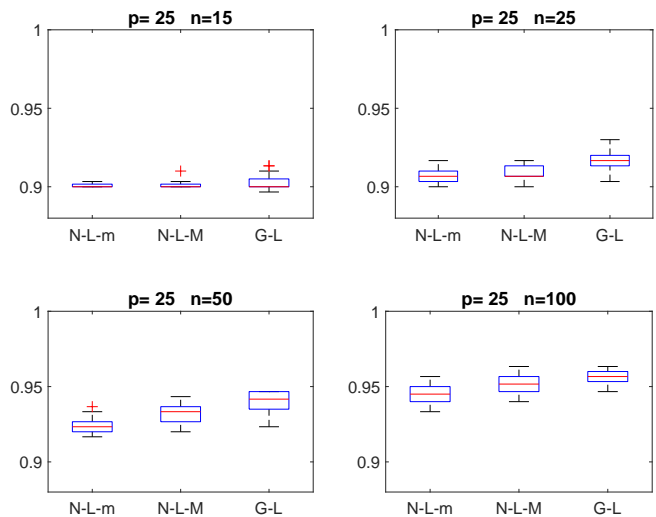}}
 \caption{ $accuracy$ box-plot obtained in 20 independent runs for the first example ($p=25$).}
\label{fig:perfG25}
\end{figure}
\begin{figure} 
\centering{
 \includegraphics[width=0.75\columnwidth]{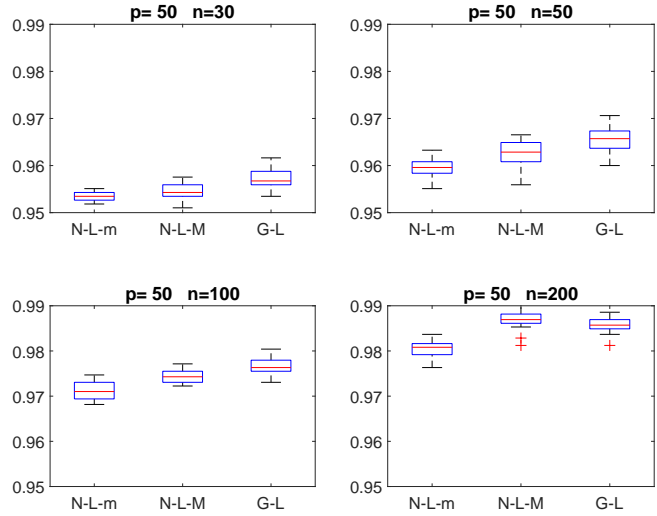}}
 \caption{  $accuracy$ Box-plot obtained in 20 independent runs for the second example ($p=50$).}
\label{fig:perfG50}
\end{figure}
\begin{figure} 
\centering{
 \includegraphics[width=0.75\columnwidth]{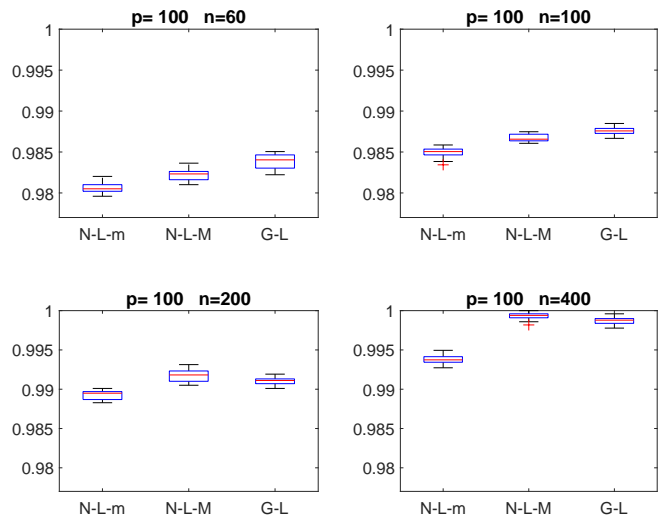}}
 \caption{  $accuracy$ Box-plot obtained in 20 independent runs for the third example ($p=100$).}
\label{fig:perfG100}
\end{figure}
\begin{figure}
\centering{
 \includegraphics[width=0.75\columnwidth]{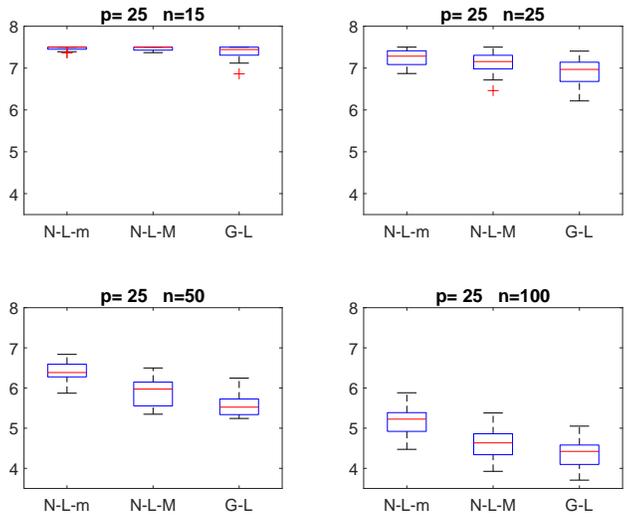}}
 \caption{ $Err$  box-plot obtained in 20 independent runs for the first example ($p=25$).}
 \label{fig:est_errorG25}
 \end{figure}
\begin{figure}
\centering{
 \includegraphics[width=0.75\columnwidth]{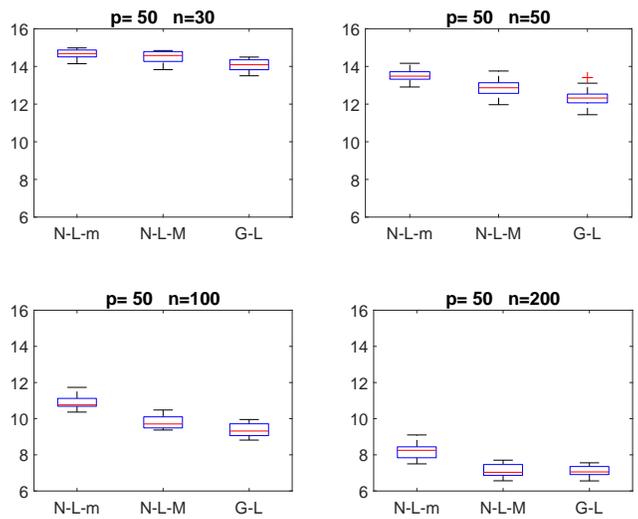}}
 \caption{  $Err$ box-plot obtained in 20 independent runs for the second example ($p=50$).}
 \label{fig:est_errorG50}
 \end{figure}

\begin{figure}
\centering{
 \includegraphics[width=0.75\columnwidth]{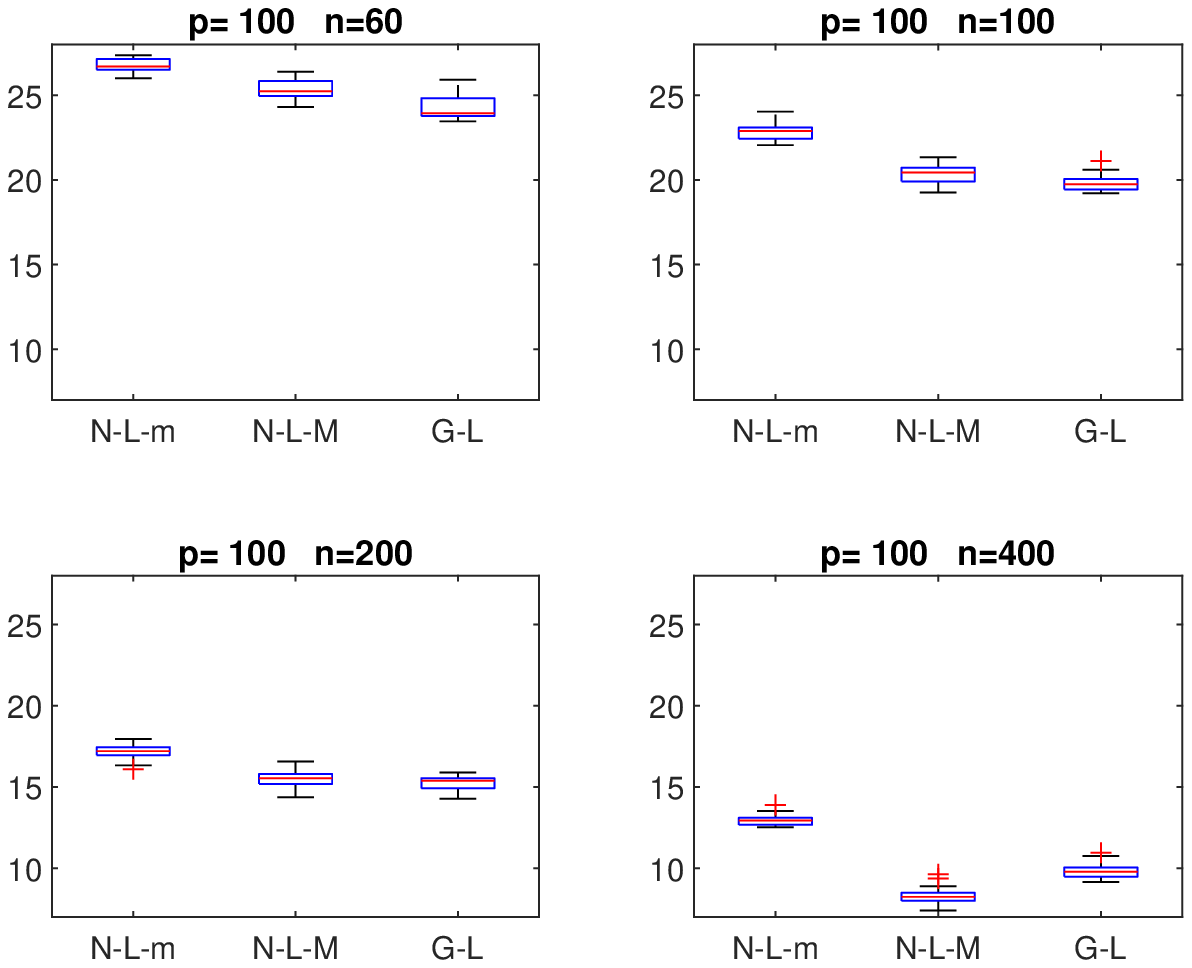}}
 \caption{  $Err$ box-plot obtained in 20 independent runs for the third example ($p=100$).}
 \label{fig:est_errorG100}
 \end{figure}

\section{Conclusions}
We presented a modification of the  node-wise logistic regression estimator, which learns the graph jointly and not one node at a time, exploiting the reciprocity of the relationship between nodes, that is the symmetry of the unknown matrix $\Theta$. Within the limits of the numerical experimentations carried out in this work, we can say that this modification slightly improves the ability of the method to reconstruct the graph but it imporves quite significantly the estimation of the link parameters. Both of these improvements are reduced as the sample size increases, decreasing the difficulty of the problem.
The proposed technique is based on the hypothesis of sparsity, which cannot be ignored. In the case of no sparsity hypothesis,  both node-wise and global approaches should be applied without the Lasso penalty, but necessarily with a number of data $n>p$ that allows the likelihood maximization without constraints.
 
Possible future developments concern the writing of an ad hoc algorithm for the proposed problem that exploits the particular form of the likelihood to accelerate its convergence. As well as, possible extension to model with interaction terms of order greater than two.

The Matlab codes used to produce results of this paper are available at http://www.iac.cnr.it/~danielad/software.html.

\section* {Acknowledgments}
This work was supported partially by CNR Italian Flagship project InterOmics and partially by INdAM-GNCS Project 2018.





\end{document}